\documentclass{article}

\PassOptionsToPackage{numbers,super}{natbib}

\usepackage[preprint]{neurips_2026}

\usepackage[utf8]{inputenc} 
\usepackage[T1]{fontenc}    
\usepackage{url}            
\usepackage{booktabs}       
\usepackage{amsfonts}       
\usepackage{nicefrac}       
\usepackage{microtype}      
\usepackage{xcolor}         

\usepackage{amsmath}
\usepackage{caption}
\usepackage{enumitem}
\setlist[enumerate]{leftmargin=2em}
\usepackage{graphicx}
\usepackage[hidelinks]{hyperref}
\usepackage{multirow}
\usepackage[section]{placeins}
\definecolor{highlightgreen}{HTML}{CBE6B8}
\definecolor{highlightblue}{HTML}{B8BFE6}
\newcommand{\highlightgreen}[1]{\setlength{\fboxsep}{0pt}\colorbox{highlightgreen}{#1}}
\newcommand{\highlightblue}[1]{\setlength{\fboxsep}{0pt}\colorbox{highlightblue}{#1}}
\newcommand{\linktext}[1]{\underline{\strut\texttt{#1}}}

\makeatletter
\def\@noticestring{
Preprint. Links:
\href{https://github.com/lintaihou/gm2}{\linktext{\texttt{code}}},\
\href{https://huggingface.co/lintaihou/gm2}{\linktext{\texttt{checkpoints}}},\ 
\href{https://api.wandb.ai/links/lintaihou/ccz3laev}{\linktext{\texttt{training curves}}}.
} 
\makeatother

\title{Graph Machine: Towards Better Pretraining via Edges}

\author{\href{https://lintaihou.com}{Lintai Hou}\\\texttt{lintai@iterlabs.ai}}

\begin{document}

\maketitle

\begin{abstract}
  We introduce the Graph Machine (GM), an architecture that maintains an $O(n)$-sized state and accesses it through sparse, dynamic routing.
  Unlike methods with fixed-size states or sparse but static routing, GM preserves $O(n)$ complexity in its sparse layers without restricting the potentially accessible state size to $O(1)$.
  Instead, GM uses edges---pointer-like objects updated differentiably by a referral mechanism resembling pointer chasing.
  We replace 75\% of the dense Transformer layers in Qwen3-0.6B with GM sparse layers and pretrain from scratch on 15.7B tokens.
  With only 2 of 4{,}096 tokens retrieved per KV head in each sparse layer, loss degrades only slightly; with 4, the best model marginally improves loss.
\end{abstract}

\vfill

\begin{figure}[!htbp]
  \centering
  \includegraphics[width=0.48\linewidth]{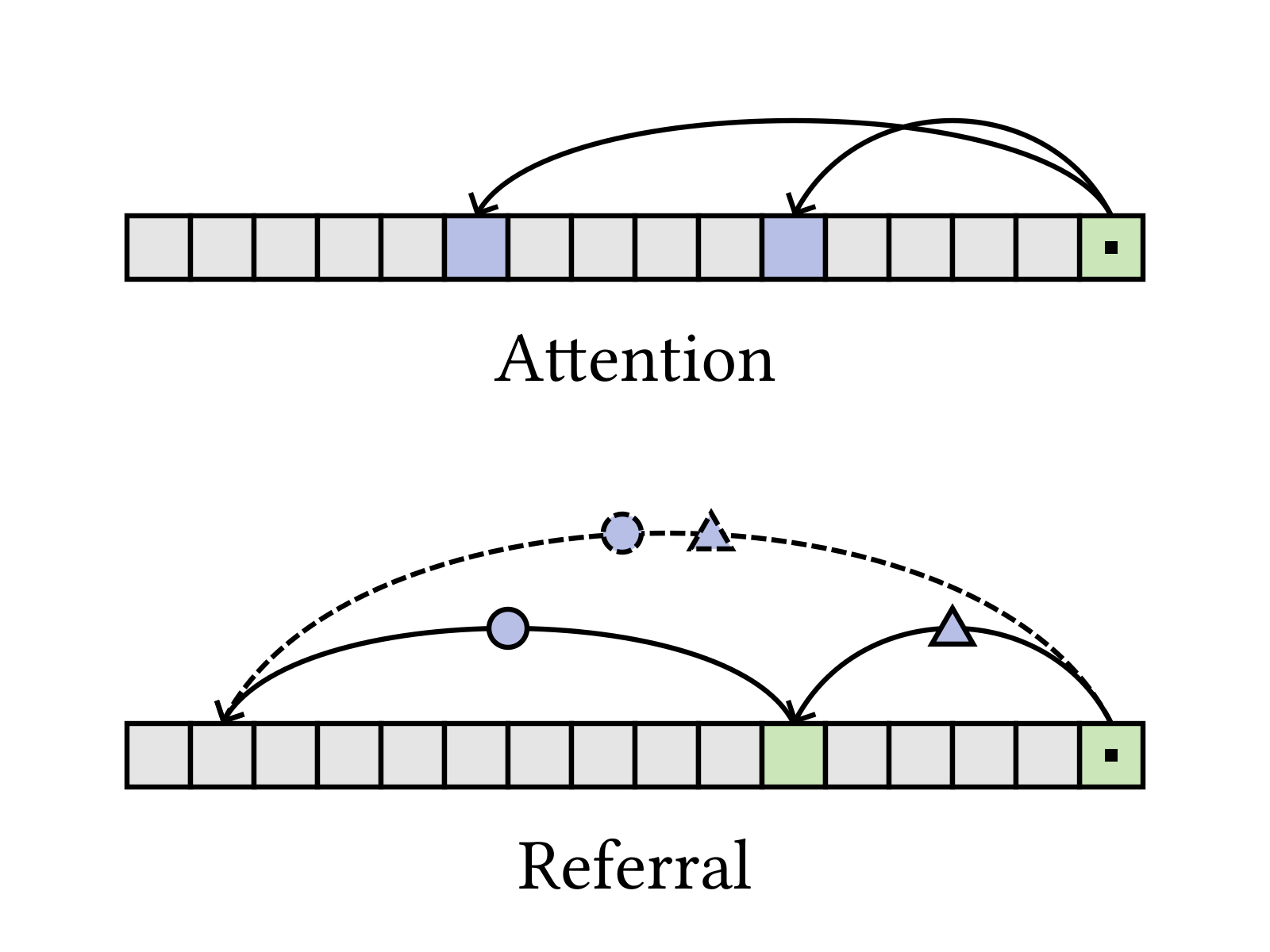}
  \caption{Attention and referral viewed as pointer operations.}
  \label{fig:pointers}
\end{figure}

\vfill

\begin{figure}[!htbp]
  \centering
  \includegraphics[width=0.48\linewidth]{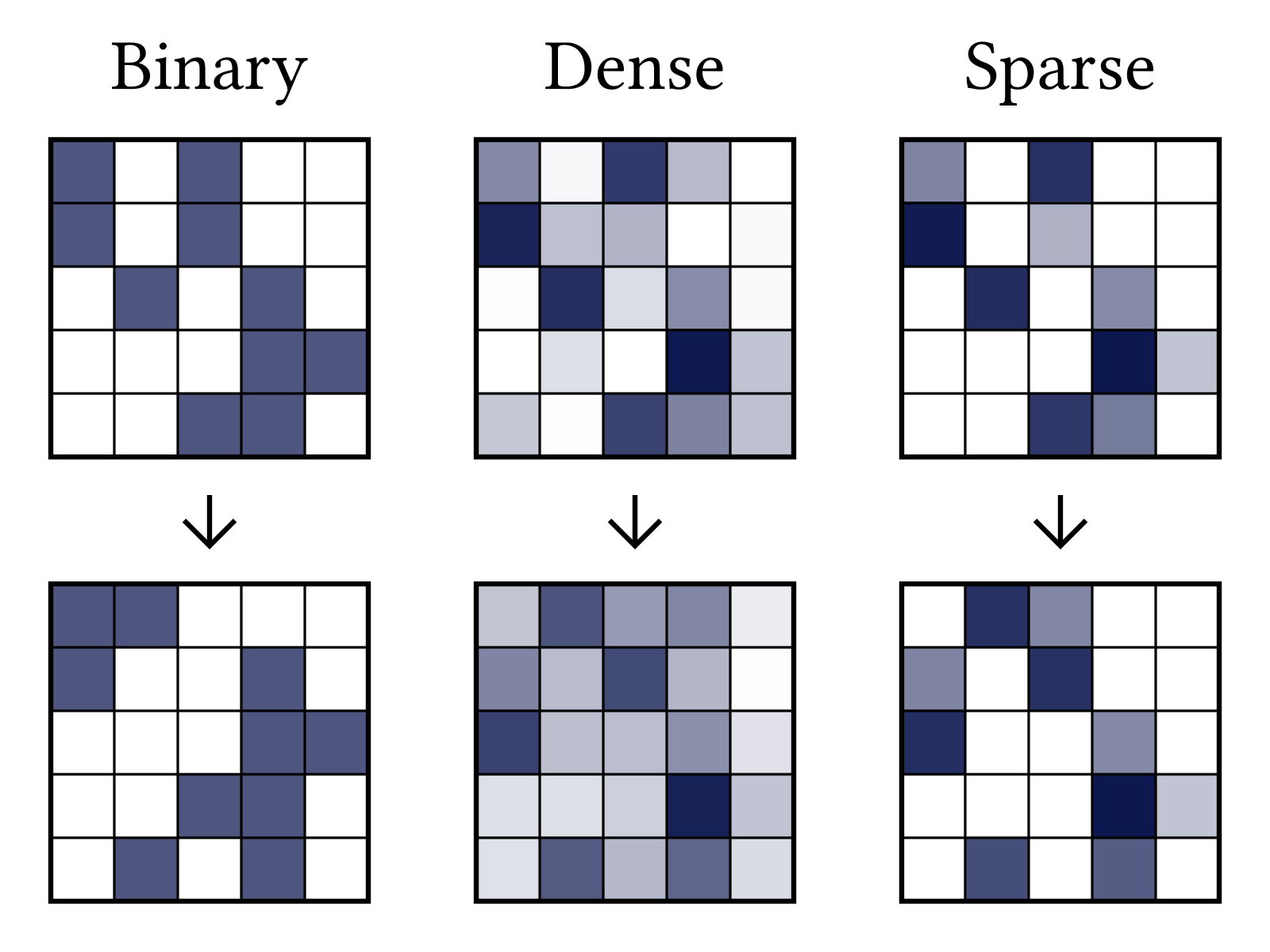}
  \caption{Different referral approaches with $\ell=2$ (simplified).}
  \label{fig:matrices}
\end{figure}

\vfill

\section{Introduction}

We classify sequence modeling approaches by the sizes of their state, access, and dynamic addresses.

Recurrent models such as RNNs~\citep{rnn} and SSMs~\citep{s4} maintain a $\Theta(1)$-sized state and must therefore compress their history.

Transformers~\citep{transformer} avoid this compression by retaining a $\Theta(n)$-sized state in their keys and values.
However, vanilla attention allows each step (token) to access the entire state, leading to $\Theta(n^2)$ complexity.

Methods such as sliding-window attention~\citep{longformer} reduce this access to $\Theta(1)$.
Yet the accessible positions are fixed independently of the new step's content: all but $\Theta(1)$ positions can be excluded in advance.
Thus, their access is sparse but static.

A simple information-theoretic argument shows that dynamic, $\Theta(1)$-sized access to a $\Theta(n)$-sized state requires each new token to provide $\Theta(\log n)$ bits
that can be used directly to address---identify and retrieve---a constant-sized portion of the state.
Together, these properties define a fourth category.

\begin{table}[!htbp]
  \caption{A sequence-model taxonomy by state size, per-step access size, and per-step dynamic address bits}
  \label{tab:taxonomy}
  \centering
  \setlength{\tabcolsep}{8pt}
  \begin{tabular}{@{\hspace{8pt}}lcccc@{\hspace{8pt}}}
    \toprule
    Type & Example(s)                 & State       & Access      & Dynamic address bits \\
    \midrule
    1    & RNN, SSM                   & $\Theta(1)$ & $\Theta(1)$ & $\backslash$         \\
    2    & Vanilla transformer        & $\Theta(n)$ & $\Theta(n)$ & $\backslash$         \\
    3    & Sliding-window transformer & $\Theta(n)$ & $\Theta(1)$ & $O(1)$               \\
    4    & GM                         & $\Theta(n)$ & $\Theta(1)$ & $\Theta(\log n)$     \\
    \bottomrule
  \end{tabular}
\end{table}

Although each dynamic address theoretically contains $\Theta(\log n)$ bits, at practical sequence lengths it fits in a fixed-width integer such as \texttt{int64}, making its storage cost effectively constant.
The Graph Machine therefore instantiates this fourth class: it maintains $\Theta(n)$ state entries, accesses $\Theta(1)$ entries per step, and uses $\Theta(\log n)$ bits of dynamic addressing per step.
Thus, GM stores both floating-point hidden states, which we call \emph{node features}, and integer edge targets which we call \emph{edge indices}.
At each new step, stored edge indices are used to retrieve $\Theta(1)$ positions for information aggregation by attention (Figure~\ref{fig:pointers}).

To update the stored edges across layers, we use a \emph{referral}~\citep{gm1} mechanism that constructs the next neighborhoods from the current $\ell$-hop neighborhoods.
Referral can be understood as neighbors recursively referring to their neighbors for $\ell$ rounds, as composing $\ell$-hop edges into one-hop edges, or as taking an adjacency matrix $A$ and producing $A^\ell$ (Figure~\ref{fig:pointers}).
If each node retains $s$ neighbors, referral can produce up to $s^\ell$ candidate paths.
Keeping the neighborhood size fixed at $s$ therefore requires a hard selection that reduces these candidates to $s$ neighbors.
To learn the selection, a probabilistic policy is needed in the form of
$$A_{\mathrm{binary}}'\sim\pi_\theta\!\left(\,\cdot\mid A_{\mathrm{binary}}^\ell\right),$$
which would introduce credit assignment difficulties.

Instead, we soften the edges rather than the policy, reformulating the top-$s$ selection into a top-$s$ approximation over soft weights.
We start with a soft and sparse adjacency matrix, take its $\ell$-th power, and then sparsify the result to form the new adjacency matrix (Figure~\ref{fig:matrices}).
I.e., we approximate $$A_{\mathrm{dense}}'=A_{\mathrm{dense}}^\ell$$ with $$A_{\mathrm{sparse}}'=\operatorname{Sparsify}_{s}\!\left(A_{\mathrm{sparse}}^\ell\right).$$
The resulting edge weights then contribute to the attention weights alongside the usual query--key factors in a product-of-experts~\citep{poe} manner.

Intuitively, referral seeks to provide useful relational representations to attention through sparsity-constrained intermediaries.
Rather than learning to shift the support directly, it learns from redistributing weight mass within the support.
From another perspective, each edge is now jointly represented by integer edge indices and floating-point \emph{edge weights}, the latter carrying gradients that discrete indices cannot.

Viewed globally, the Graph Machine maintains and updates a graph with soft directed edges.
Sparse attention passes features along edges to update nodes, while referral passes addresses along edges to form new ones.
In this paper, we hybridize GM and Transformer layers and train the resulting Graph Language Machines (GLMs) under a standard language-model pretraining setup.
Throughout the paper, we move between the vocabularies of adjacency matrices, edges, neighbors, addresses, pointers, and indices as appropriate.

\section{Architecture}
We first provide an overview of the GM architecture, then introduce two common operations, and finally describe its two new sparse edge submodules.

\subsection{Overview}

\begin{figure}[!htbp]
  \begin{minipage}{0.48\textwidth}
    \centering
    \includegraphics[width=\linewidth]{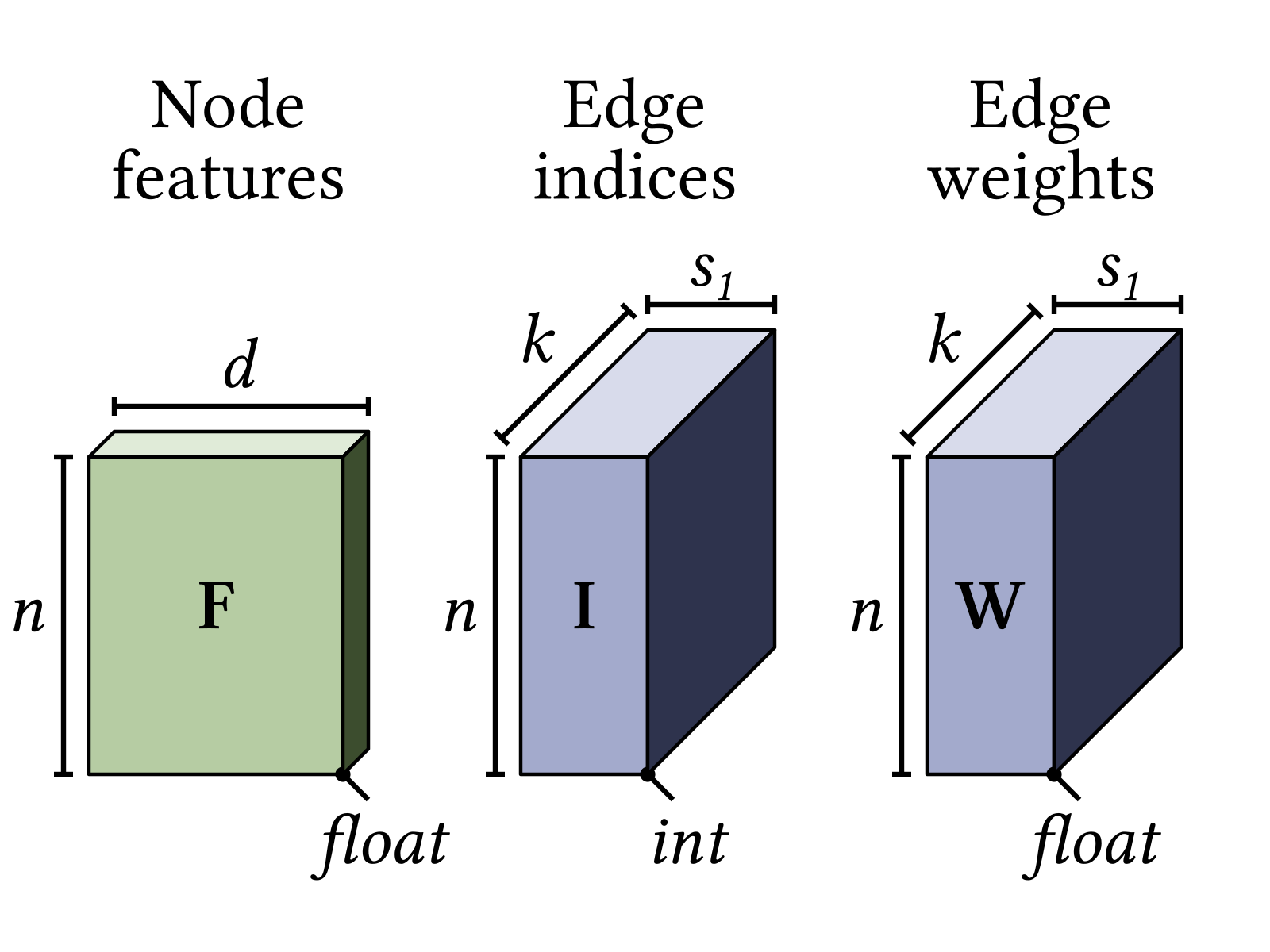}
    \caption{GM states.}
    \label{fig:states}
  \end{minipage}
  \hfill
  \begin{minipage}{0.48\textwidth}
    \centering
    \includegraphics[width=\linewidth]{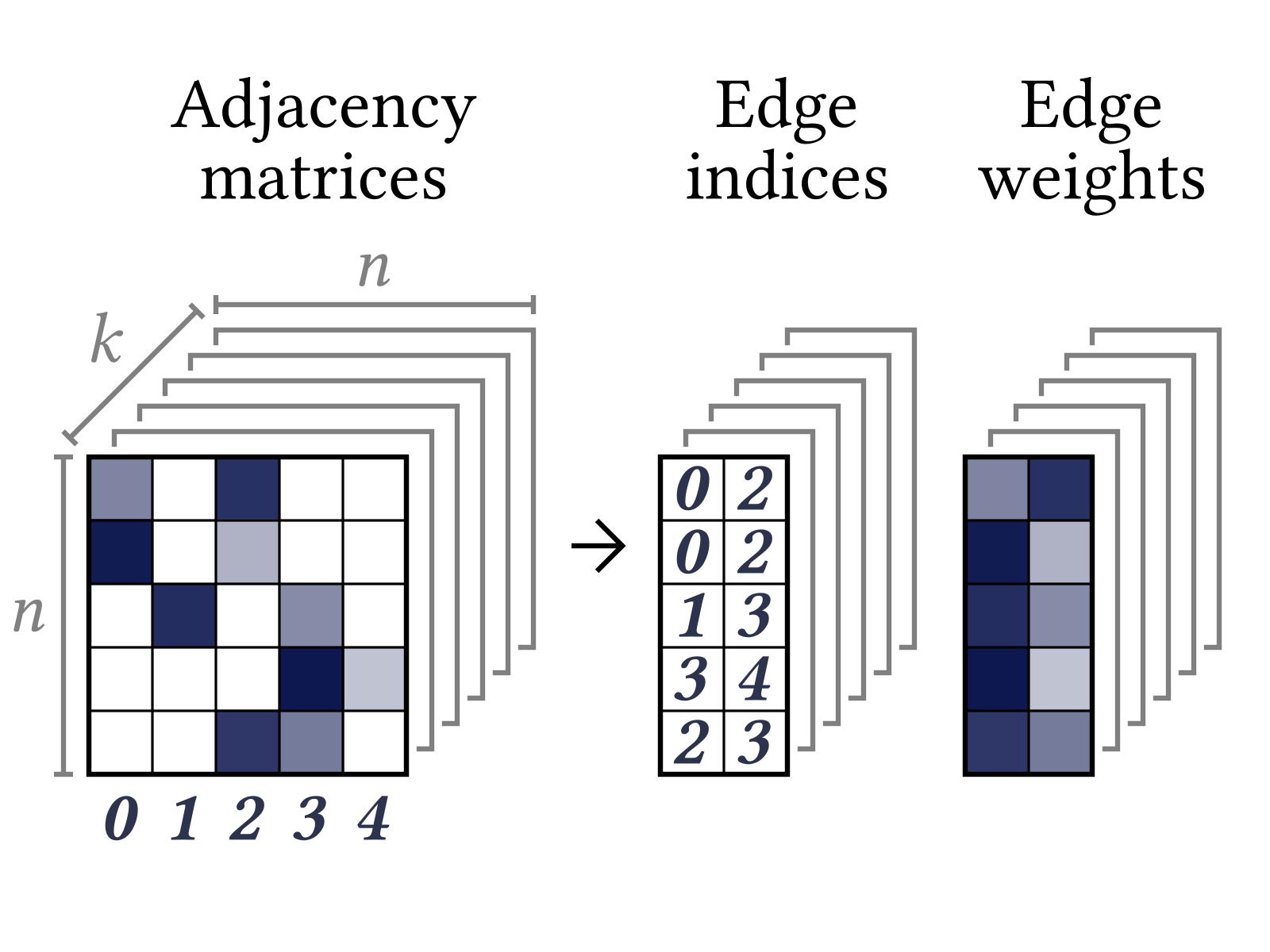}
    \caption{Adjacency matrices → edge states.}
    \label{fig:ell}
  \end{minipage}
\end{figure}

\paragraph{States.}
GM maintains three state tensors (Figure~\ref{fig:states}).
Alongside the $n \times d$ node features, it maintains edge indices and edge weights, both of shape $n \times k \times s_1$.
We obtain this representation by generalizing the adjacency matrix to $k$ copies and representing them in an $s_1$-sparse coordinate format, giving each node $k$ edges and each edge $s_1$ member positions (Figure~\ref{fig:ell}).
Across the $s_1$ member positions, the edge-index tensor specifies target nodes, while the edge-weight tensor provides corresponding nonnegative weights that sum to one.
We call the edge states passed between layers stored edges, distinguishing them from the mixed edges used directly by referral and attention within the submodules.

\begin{figure}[!htbp]
  \begin{minipage}{0.48\textwidth}
    \centering
    \includegraphics[width=\linewidth]{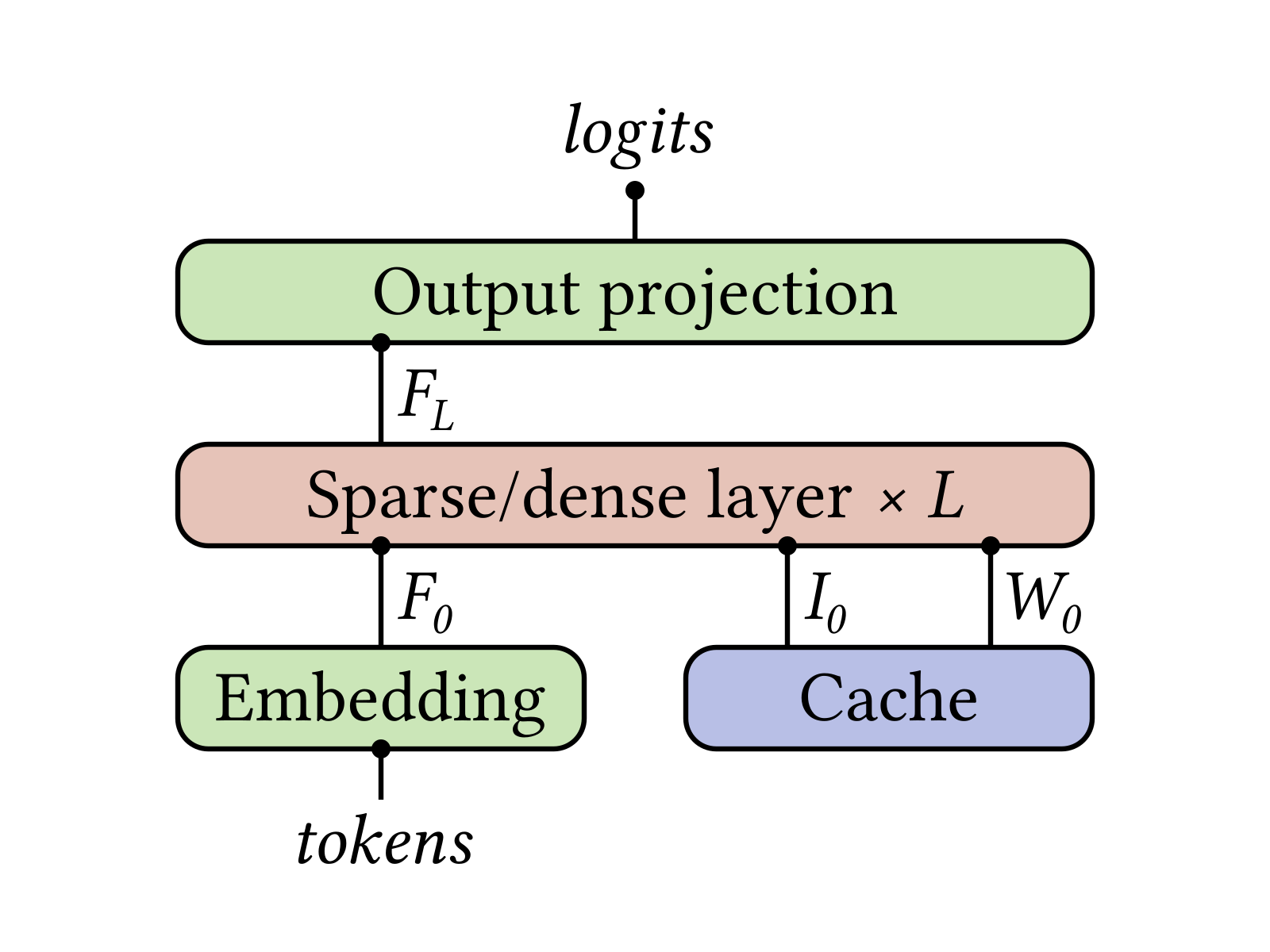}
    \caption{GM layers.}
    \label{fig:layers}
  \end{minipage}
  \hfill
  \begin{minipage}{0.48\textwidth}
    \centering
    \includegraphics[width=\linewidth]{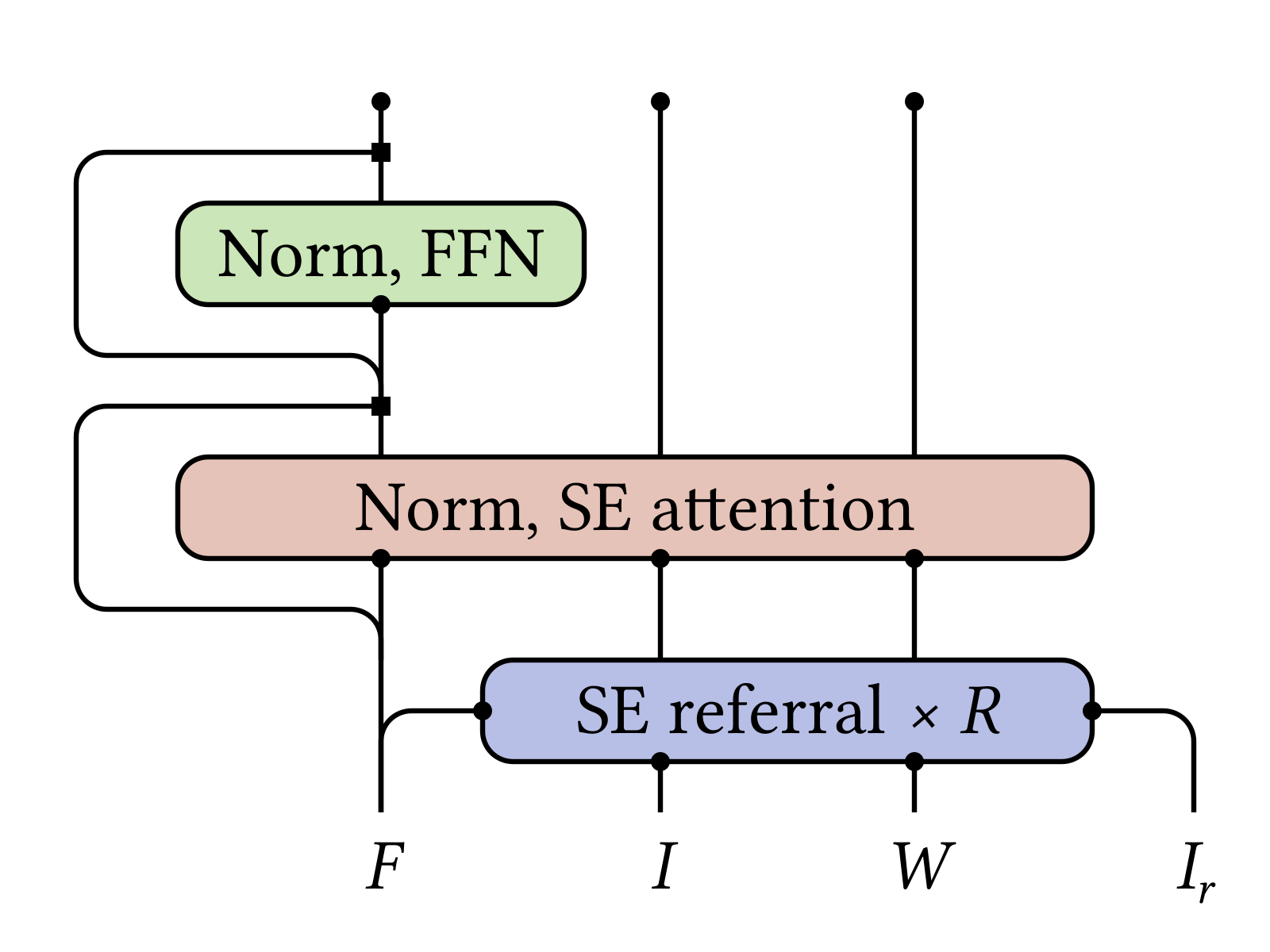}
    \caption{A sparse GM layer.}
    \label{fig:sparse-layer}
  \end{minipage}
\end{figure}

\paragraph{Layers.}
Token embeddings initialize the node features in our GLMs.
Each token's initial input edges point to itself and its $k-1$ nearest preceding tokens, clamped to the first token and cached for reuse.
Each input edge places all its weight on its single target and fills its remaining member positions with zero indices and weights.
The dense layers are standard Transformer layers.
Each sparse layer applies $r$ edge-referral submodules, each performing two-hop referral, followed by an edge-attention submodule and an MLP.
Referral updates the edges, while attention and the MLP update the node features.
Edges remain causal because they are derived from previous causal edges or masked attention.
As usual, both dense and sparse layers use pre-normalization and residual connections for node features.
A final output projection produces prediction logits from node features.

\subsection{Operations}

\paragraph{Sparsify.}
A conceptual weighted average of adjacency-matrix rows requires several additional steps in a sparse-coordinate implementation.
Duplicate indices must be coalesced, the highest-weight entries selected, and their weights renormalized to achieve the target sparsity.
For example, equally averaging $([0,3],[0.4,0.6])$ and $([2,3],[0.8,0.2])$ first produces $$([0,3,2,3],[0.2,0.3,0.4,0.1]).$$
Coalescing the two entries for index $3$, dropping the lowest-weight entry, and renormalizing gives $$([2,3],[0.5,0.5]).$$
Since we use hard top-$s$ selection, only the retained weights receive gradients.
Sparsification occurs when stored edges are mixed before referral and attention, and when the resulting edges are reconciled after referral.

\paragraph{Mix.}
We mix the $k$ stored edges of each node before using them for referral or attention, much as a pointwise convolution mixes channels in a CNN.
For each output edge, mixing logits are projected from the node features and normalized over the input edges with a softmax, producing a mixing matrix $M$.
The resulting coefficients are used to compute a weighted average of the input edges.
Weighted averaging $\widetilde{A}=MA$ alone would keep new edges approximately within the convex hull of earlier edges.
We therefore introduce nonlinearity through temperature scaling both before and after the average.
Separate input- and output-edge temperatures are parameterized as the softplus or exponential of projections from the node features; each acts as an exponent on the corresponding edge weights, which are then renormalized.
I.e., for an edge-weight vector $a$, the temperature-scaling is $$\operatorname{T}_{\tau}(a)=\frac{a^\tau}{\sum_j a_j^\tau},\quad\tau>0,$$
and mixing follows $$\widetilde{A}=\operatorname{T}_{\boldsymbol{\tau}_{\mathrm{out}}}\left(M\,\operatorname{T}_{\boldsymbol{\tau}_{\mathrm{in}}}(A)\right).$$
Sparsification coalesces the mixed edges $\widetilde{A}$ and reduces them to the desired sparsity.

\subsection{Submodules}

\paragraph{Sparse edge referral (SER).}
During sparse edge referral, rather than using a single adjacency matrix $A$, we take two matrices $A_1$ and $A_2$ and perform two-hop referral via
$$A'=\operatorname{Sparsify}_{s_1}\!\left(A_1A_2\right).$$
At a path level, referral composes a first edge $e_1=(n_1\xrightarrow{w_1}n_2)$ with a second edge $e_2=(n_2\xrightarrow{w_2}n_3)$, omitting sparsification:
$$e_1\circ e_2=\left(n_1 \xrightarrow{w_1w_2} n_3\right).$$
Concretely, to construct $k$ new edges, we first mix the $k$ stored edges into $2k$ channels, forming $k$ pairs that represent the two legs of the new edges.
Within each pair, $e_1$ and $e_2$ are sparsified to $s_2$ and $s_3$, respectively.
The $e_1$ indices are then used to retrieve the corresponding $e_2$ indices and weights.
This produces up to $s_2s_3$ members per edge before sparsification to $s_1$.
Optionally, refresh edges augment the stored edges before mixing.
These refresh edges can come from the initial input edges or from the top-$k$ indices and normalized weights of the most recent dense-attention layer.

\paragraph{Sparse edge attention (SEA).}
During sparse edge attention, we mix the $k$ stored edges into $g$ channels, one per KV head, and sparsify each to $s_4$.
The resulting edge indices specify which positions provide the keys and values.
We compute the usual scaled query--key scores over these positions and combine them with the mixed edge weights as a product of experts.
Specifically, we call the query--key scores \emph{node factors} and scale them using per-head temperatures parameterized analogously to the output-edge temperatures in mixing.
We call the logarithms of the mixed edge weights \emph{edge factors}.
Adding the node and edge factors and applying softmax produces the final attention weights.
Formally, for one query head and its associated KV head, let $\mathcal{N}(i)$ denote the positions retrieved for node $i$, and let $w_{ij}$ denote their mixed edge weights.
The final attention weights are $$a_{ij}=\operatorname{softmax}_{j\in\mathcal{N}(i)}\left(\tau\frac{q_i^\top k_j}{\sqrt{d_h}}+\log w_{ij}\right)\propto w_{ij}\exp\left(\tau\frac{q_i^\top k_j}{\sqrt{d_h}}\right).$$
Here, addition in logit space corresponds to a product of experts in probability space.
Value aggregation and output projection then follow as in standard attention.
Optionally, we use the retrieved indices and final attention weights to replace a subset of the stored edges.
This realigns the pointer distribution after approximate referral using the feature-based evidence obtained during attention.

\section{Results}

We use Qwen3-0.6B~\citep{qwen3} as our baseline and as a representative modern dense LLM.
For a controlled comparison, all models are trained from scratch using the same backbone and training hyperparameters, codebase, and random seed.
We use a conventional LLM training recipe based on established practice, without tuning it to our specific conditions.

For GM, we use a sparse-to-dense layer ratio of $3{:}1$, scheduled as $[\mathrm{S},\mathrm{S},\mathrm{D},\mathrm{S}] \times 7$.
We use 16--32 edges, 0--6 referral steps, sparsity budgets of $(2,4,4,2)$ or $(4,4,4,4)$, 8 input-refresh edges, and 0--8 dense-refresh edges, with realignment enabled.
Temperatures are parameterized by the horizontally shifted softplus $T(x)=\operatorname{softplus}(x+\log(e-1))$, such that $T(0)=1$.
RoPE~\citep{rope} is omitted from SEA.

We name each GLM by its sparsity budget, number of stored edges, number of referral steps, and whether it uses eight dense-refresh edges.
Theia~\citep{007} models use a sparsity budget of $(2,4,4,2)$, whereas Hyperion models use $(4,4,4,4)$.
Because $s_4$ determines attention sparsity, Theia and Hyperion retrieve 2 and 4 positions per KV head during SEA, respectively.
For example, Hyperion-K16-R3-S uses a $(4,4,4,4)$ sparsity budget, 16 edges, three referral steps, and eight dense-refresh edges.
In the tables, we group conditions by sparsity and then order them by the number of referral steps.

\begin{table}[!htbp]
  \caption{Parameter counts and estimated training FLOPs.
    Only multiplications and accumulations are counted, each as one FLOP; normalization, softmax, activations, RoPE, and sparsification are excluded.
    SEA KV access is measured over a 4,096-token sequence relative to dense causal attention.}
  \label{tab:params}
  \centering
  \setlength{\tabcolsep}{4pt}
  \begin{tabular}{@{\hspace{8pt}}lcccc@{\hspace{8pt}}}
    \toprule
    Model             & Parameters & Total compute & Ref + att compute & SEA att KV access   \\
    \midrule
    Qwen3             & 596M       & 78.4 EFLOPs   & 38.8 EFLOPs       & 100.000\%           \\
    \midrule
    Theia-K24         & 601M       & 62.3 EFLOPs   & 22.7 EFLOPs       & \phantom{00}0.098\% \\
    Theia-K32-R2      & 711M       & 72.7 EFLOPs   & 33.0 EFLOPs       & \phantom{00}0.098\% \\
    Theia-K24-R3      & 700M       & 71.6 EFLOPs   & 32.0 EFLOPs       & \phantom{00}0.098\% \\
    Theia-K24-R4      & 733M       & 74.7 EFLOPs   & 35.1 EFLOPs       & \phantom{00}0.098\% \\
    Theia-K16-R4-S    & 686M       & 73.1 EFLOPs   & 33.5 EFLOPs       & \phantom{00}0.098\% \\
    Theia-K16-R6      & 700M       & 71.6 EFLOPs   & 32.0 EFLOPs       & \phantom{00}0.098\% \\
    \midrule
    Hyperion-K32-R1   & 657M       & 67.5 EFLOPs   & 27.9 EFLOPs       & \phantom{00}0.195\% \\
    Hyperion-K24-R3   & 700M       & 71.6 EFLOPs   & 32.0 EFLOPs       & \phantom{00}0.195\% \\
    Hyperion-K16-R3   & 650M       & 66.9 EFLOPs   & 27.3 EFLOPs       & \phantom{00}0.195\% \\
    Hyperion-K16-R3-S & 664M       & 71.1 EFLOPs   & 31.4 EFLOPs       & \phantom{00}0.195\% \\
    Hyperion-K16-R4   & 666M       & 68.5 EFLOPs   & 28.8 EFLOPs       & \phantom{00}0.195\% \\
    \bottomrule
  \end{tabular}
\end{table}

We train on a randomly sampled, overprovisioned subset of FineWeb-Edu~\citep{fineweb}, which is processed for less than one epoch.
Documents are packed into 4,096-token sequences and padded as needed.
A padding token is placed at the start of each sequence to absorb underflowing edge indices.
Across a random sample of 100K documents, document length has a mean of 1,035 and a standard deviation of 1,909 Qwen3 tokens.
A sequence length of 4,096 therefore does not imply a comparable effective context length within each document.

The implementation is primarily written in generic PyTorch~\citep{pytorch}, with a custom Triton~\citep{triton} kernel used only for sparsification.
Each training run uses a single H100 SXM and takes 53--236 hours, with Qwen3 requiring 53 hours and most GLMs clustering around 150--160 hours.
Under our prototype implementation, GLMs are generally several times slower than Qwen3 on this hardware.
Preliminary experiments on an RTX 4090 show that some configurations approach Qwen3's training throughput, suggesting that relative performance depends strongly on hardware and kernel implementation.

With the backbone dimensions held fixed, the referral and temperature parameters increase the parameter counts of referral-equipped GLMs by $9\%$--$23\%$.
Nevertheless, these models reduce total estimated training compute by $5\%$--$15\%$ and referral-plus-attention compute by $10\%$--$30\%$ relative to Qwen3.
Most of the additional parameters belong to referral projections, which support relatively inexpensive operations.
Over a length-$n$ sequence, dense causal attention accesses an average of $(n+1)/2$ KV positions per query.
At $n=4096$, retrieving at most two or four positions therefore corresponds to $0.098\%$ or $0.195\%$ of dense causal KV access, respectively.

\begin{table}[!htbp]
  \caption{Test loss throughout training.
    Qwen3 entries give absolute loss, whereas GLM entries give the difference from Qwen3.
    Highlighting marks the model with the lowest final loss in each sparsity category.}
  \label{tab:results}
  \centering
  \setlength{\tabcolsep}{16pt}
  \begin{tabular}{@{\hspace{8pt}}lcccc@{\hspace{8pt}}}
    \toprule
    \multirow{2}{*}[-2.5pt]{Model} & \multicolumn{4}{c}{Test loss / $\Delta$ test loss}                                                                               \\
    \cmidrule(lr){2-5}
                                   & @ 25\%                                             & @ 50\%                  & @ 75\%                  & @ 100\%                 \\
    \midrule
    Qwen3                          & \phantom{+}2.869                                   & \phantom{+}2.744        & \phantom{+}2.660        & \phantom{+}2.587        \\
    \midrule
    Theia-K24                      & +0.041                                             & +0.039                  & +0.040                  & +0.040                  \\
    Theia-K32-R2                   & +0.016                                             & +0.017                  & +0.018                  & +0.020                  \\
    Theia-K24-R3                   & \highlightgreen{+0.012}                            & \highlightgreen{+0.012} & \highlightgreen{+0.013} & \highlightgreen{+0.014} \\
    Theia-K24-R4                   & +0.013                                             & +0.013                  & +0.014                  & +0.015                  \\
    Theia-K16-R4-S                 & +0.010                                             & +0.013                  & +0.014                  & +0.016                  \\
    Theia-K16-R6                   & +0.013                                             & +0.014                  & +0.014                  & +0.015                  \\
    \midrule
    Hyperion-K32-R1                & +0.008                                             & +0.008                  & +0.009                  & +0.009                  \\
    Hyperion-K24-R3                & -0.002                                             & -0.003                  & -0.003                  & -0.002                  \\
    Hyperion-K16-R3                & \phantom{+}0.000                                   & \phantom{+}0.000        & -0.001                  & \phantom{+}0.000        \\
    Hyperion-K16-R3-S              & \highlightblue{-0.005}                             & \highlightblue{-0.005}  & \highlightblue{-0.005}  & \highlightblue{-0.003}  \\
    Hyperion-K16-R4                & -0.004                                             & -0.003                  & -0.003                  & -0.002                  \\
    \bottomrule
  \end{tabular}
\end{table}

All models are evaluated on the same fixed held-out random subset of FineWeb-Edu.
Final test losses are around 2.60, consistent with broader LLM pretraining experience at this scale.

Increasing the number of edges beyond the 16 attention heads provides a moderate improvement in the matched Hyperion comparison: K24-R3 outperforms K16-R3 by $0.003$ at the end of training.
This improvement comes at additional parameter and compute cost because full cross-edge mixing scales as $k^2$.

Referral clearly improves performance from zero to a few steps: Theia-K24-R3 improves over the non-referral Theia-K24 baseline by $0.026$ at $100\%$.
Additional referral steps can also help under some conditions, with Hyperion-K16-R4 improving over Hyperion-K16-R3 by $0.003$.
However, this trend does not hold in every setting.
Under the sparser Theia budget with more edges than heads, Theia-K24-R4 performs $0.001$ worse than Theia-K24-R3.

Preliminary experiments suggest benefits from realignment and input refresh.
We observe a similar benefit from dense refresh: Hyperion-K16-R3-S improves over Hyperion-K16-R3 by $0.004$ at $100\%$.

Within the tested configurations, neither parameter count nor estimated compute is a strong predictor of performance.
Hyperion-K16-R3-S achieves the best loss despite being among the less expensive GLMs, with $11\%$ more parameters and $19\%$ less referral-plus-attention compute than Qwen3.

Overall, the best Theia model increases final loss by approximately $0.014$, while the best Hyperion model reduces it by approximately $0.003$.
These results show that most Transformer layers can be replaced by sparse layers retrieving only 2 or 4 of 4,096 positions per KV head without materially sacrificing quality, while meaningfully reducing estimated compute.

\section{Related work}

This work builds directly on the original GM work~\citep{gm1}; we refer to the two papers as GM-1 and GM-2.
\begin{enumerate}
  \item GM-1 uses a bespoke Sudoku benchmark, whereas GM-2 adopts a standard language-model pretraining setting, making its results easier to contextualize.
  \item GM-1 primarily studies the inductive bias introduced by edges, leaving computational practicality to future work.
        Its dense edge representation incurs cubic time and quadratic space complexity, limiting experiments to a few hundred nodes.
        GM-2 instead uses sparse edge representations and operations.
        Its edge indices and weights form a sparse coordinate representation of GM-1's edge addresses, while SER and SEA are sparse counterparts of GM-1's referral and attention operations.
  \item GM-2 simplifies the state by unifying node and edge features, removing the need for carefully engineered interactions among separate representations.
  \item GM-2 introduces refresh and realignment, allowing referral to reuse initial or dense-attention-derived edges and sparse attention to update stored edges.
\end{enumerate}

GM is related to efficient sequence models and their hybrids~\citep{mamba,gdn,jamba,kda}.
GM is also situated within the sparse-attention literature~\citep{sparse-transformer,longformer,routing-transformer,nsa,dsa}.

\section{Limitations and conclusion}

This work demonstrates the GM architecture while leaving substantial room for architectural and implementation improvements.
More efficient custom kernels are one such important direction.

Our experiments are small-scale in both model and training, and evaluate only a pretraining setup using test loss as an aggregate metric.
We leave richer evaluations across scales, setups, and downstream capabilities to future work.

Although inductive bias was a central focus of GM-1, it remains to be examined for the updated architecture and in the context of language modeling.
As part of this investigation, but also as an independent direction, mechanistic interpretation could take advantage of GM's highly self-interpretable relational states and operations.

In conclusion, although some amount of global dense attention likely remains valuable under current language-modeling settings, our results show---at least at our scale, under our setup, and by our measure---that it is nonetheless highly trimmable.
If part of attention's role is fundamentally to pass and resolve addresses, then we can make the core machinery---logarithmic-sized identification bits and direct retrieval through them---primitives of the architecture.
GM now carries and transmits addresses via indices rather than features, and resolves them through indexed gathering rather than dense scoring.

This creates new degrees of freedom for architectural design.
On the efficiency side, dense computation can be traded off against sparse memory operations.
On the inductive-bias side, architectures can trade off relational traversal against global search, and address passing against content passing.
Finding the optimal balance will require further exploration.

\clearpage

\bibliographystyle{unsrt}
\bibliography{graph-machine-2}

\end{document}